\documentclass[letterpaper, 10pt, conference]{ieeeconf}  

\IEEEoverridecommandlockouts                              

\usepackage{amsmath,amsfonts}
\usepackage{multirow}
\usepackage{graphicx}
\usepackage{subfigure}
\usepackage{algorithm}
\usepackage[noend]{algorithmic}

\usepackage{caption}
\usepackage{cite}

\title{\LARGE \bf
Improving Offline Reinforcement Learning with Inaccurate Simulators
}

\author{Yiwen Hou, Haoyuan Sun, Jinming Ma, and Feng Wu$^*$
\thanks{This work was supported in part by the Major Research Plan of the National Natural Science Foundation of China (92048301), Anhui Provincial Major Research and Development Plan (202004H07020008), and Anhui Province Development and Reform Commission 2021 New Energy and Intelligent Connected Vehicle Innovation Project.}
\thanks{All authors are with the School of Computer Science and Technology, University of Science and Technology of China, Hefei, Anhui, China.
E-mail: (\{houyiwen,sunhaoyuan,jinmingm\}@mail.ustc.edu.cn, wufeng02@ustc.edu.cn). $^*$Feng Wu is the corresponding author.}
}

\begin{document}

\maketitle
\thispagestyle{empty}
\pagestyle{empty}

\begin{abstract}
Offline reinforcement learning (RL) provides a promising approach to avoid costly online interaction with the real environment. However, the performance of offline RL highly depends on the quality of the datasets, which may cause extrapolation error in the learning process. In many robotic applications, an inaccurate simulator is often available. However, the data directly collected from the inaccurate simulator cannot be directly used in offline RL due to the well-known exploration-exploitation dilemma and the dynamic gap between inaccurate simulation and the real environment. To address these issues, we propose a novel approach to combine the offline dataset and the inaccurate simulation data in a better manner. Specifically, we pre-train a generative adversarial network (GAN) model to fit the state distribution of the offline dataset. Given this, we collect data from the inaccurate simulator starting from the distribution provided by the generator and reweight the simulated data using the discriminator. Our experimental results in the D4RL benchmark and a real-world manipulation task confirm that our method can benefit more from both inaccurate simulator and limited offline datasets to achieve better performance than the state-of-the-art methods.

\end{abstract}

\section{Introduction}

\noindent Deep reinforcement learning (RL) has shown impressive success in many robotic applications \cite{gu2017deep}. However, applying RL to real-world scenarios is still very challenging because exploration and interaction with real-world environments are often costly or risky for physical robots, and RL methods often require millions of such data to learn a good policy. 
Most recently, offline RL emerges as a promising solution to address the dilemma above, allowing for learning efficient policies offline entirely from previously collected data \cite{levine2020offline}. 



However, Offline RL presents several significant challenges, such as the extrapolation error incurred by the mismatch between the experience distributions of the learned policy and the dataset \cite{fujimoto2019off}. To minimize this error, most prior work attempts to constrain the trained policy to the offline dataset’s action space \cite{fujimoto2019off, kumar2019stabilizing, wu2019behavior, fujimoto2021minimalist},  value regularization \cite{kumar2020conservative, yu2021combo} on out-of-distribution (OOD) actions, or weighted \cite{kostrikov2021offline, peng2019advantage, brandfonbrener2021offline, xu2023offline} or conditioned \cite{xu2022policy, emmons2021rvs, yang2021rethinking} behavior cloning. 
Although these methods achieve considerable success in the offline setting, they are still heavily reliant on the quality of the offline dataset \cite{arnob2021importance, chen2019information, schweighofer2022dataset}. If a large portion of state-action space is not explored within the dataset, offline RL methods usually fail to learn good policies. In real-world scenarios, data collection is commonly expensive or limited, resulting in suboptimal and noisy offline datasets. This directly limits the potential application of offline RL in robotics.

Fortunately, in many robotics tasks, a simulator (e.g. Gazebo \cite{koenig2004design}, SimSpark \cite{xu2014simspark}, MuJoCo \cite{todorov2012mujoco}) is often available. Within a simulator, robots can engage in unrestricted exploration, gaining access to a vast array of state-action data. These data can compensate for the limitations of the offline dataset in hand. Although the combination of real-world data and simulated data seems promising, there are two major challenges within this approach. 
Firstly, although the simulator provides a risk-free environment for exploration, low-quality or aimless exploration data might not be sufficient and effective in addressing the OOD issues within the offline dataset. 
Thus, striking a balance between the {\em exploration} in the simulator and the {\em exploitation} of offline datasets becomes crucial.
Secondly, accurately modeling the complex dynamics of the real world is often intractable or expensive. In contrast, {\em inaccurate simulators} are relatively easier to obtain and more efficient (e.g., MuJoCo for RL). However, if we indiscriminately treat the inaccurate simulated data equally with the real offline data, the {\em dynamics gap} \cite{niu2022trust} between the real and simulated environments might adversely impact the effectiveness of offline policy learning.

\begin{figure}[t] 
    \centering
    \includegraphics[width=0.85\columnwidth]{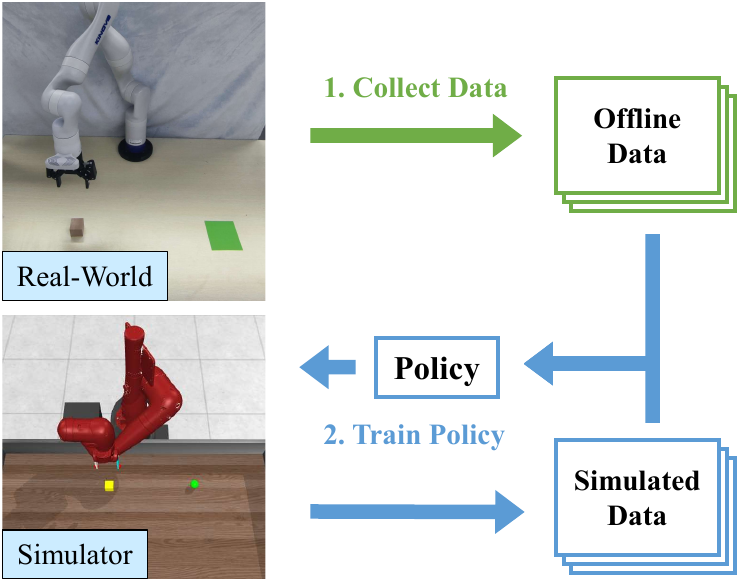}
    \caption{Combination of offline datasets from real world and an inaccurate simulator (MuJoCo) for improving offline RL.}
    \label{fig:data_distribution}
\end{figure}

To address these issues, we propose a novel method called \textbf{O}ffline \textbf{R}einforcement learning with \textbf{I}naccurate \textbf{S}imulator (ORIS), as shown in Fig. \ref{fig:data_distribution}, which aims to 1) collect more effective data from the inaccurate simulator to achieve exploration-exploitation trade-off, and 2) better use the mixed datasets to alleviate the side-effect of inaccurate simulation. 
To this end, we first pre-train a Generative Adversarial Network (GAN) \cite{goodfellow2020generative} model with the offline dataset to fit the state distribution of the offline dataset. Given this, we interact with an inaccurate simulator by a hybrid rollout policy from the initial state distribution provided by the GAN generator, which effectively balances the exploration-exploitation dilemma.
Then, we employ the GAN discriminator to adeptly integrate both the offline datasets and the simulated data, reducing the side-effect of inaccurate simulator and enhancing the precision of Q value estimations.
We conducted experiments on the D4RL dataset \cite{fu2020d4rl}, a common offline RL benchmark, as well as real-world robotic manipulations. The experimental results show that our method achieved better performance than the state-of-the-art methods especially given limited amount of data.

\section{Background}
\noindent We formally model the robotic RL problem as a Markov Decision Process (MDP): $\mathcal{M} = (\mathcal{S, A, P}, \rho_0, r, \gamma )$, where $\mathcal{S}$ is the state space, $\mathcal{A}$ is the action space; $\mathcal{P}: \mathcal{S} \times \mathcal{A} \times \mathcal{S} \rightarrow [0, 1]$ is the environment dynamics; $\rho_0\in \Delta(S)$ is the distribution of the initial states;  $r(s, a): \mathcal{S \times A} \rightarrow \mathbb{R} $ is the reward function; $\gamma \in (0, 1]$ is the discount factor. The goal of RL is to learn a policy $\pi(a | s): \mathcal{S} \rightarrow \Delta(\mathcal{A})$ that maximizes the cumulative discounted returns: $\sum_{t=0}^{\infty}\gamma^t r(s_t, a_t)$ from the experiences without directly accessing the model.



To date, off-policy actor-critic algorithms are one of the most commonly used frameworks to solve the RL problem without consideration of how the experiences were generated, which learn a Q-function $ Q_{\theta}(s, a)$ by minimizing the Bellman error and a policy $\pi_{\phi}$ by maximizing the Q-function, where $\theta$ and $\phi$ are the parameters of Q-function and policy, and the loss functions are as follow:
\begin{eqnarray}
    \label{critic loss}
    \mathcal{L}_{Q}(\theta) &=& \mathop{\mathbb{E}}_{(s, a, s^{\prime}) \sim \mathcal{B}} \left[ (Q_{\theta}(s, a) - \mathcal{T^{\pi_{\phi}}}Q_{\theta}(s, a) )^{2} \right] \\
    \label{actor loss}
    \mathcal{L}_{\pi}(\phi) &=& \mathop{\mathbb{E}}_{(s) \sim \mathcal{B}, a \sim \pi_{\phi}(\cdot|s)} \left[- Q_{\theta}(s, a)\right]
\end{eqnarray}
where $\mathcal{B}$ is the replay buffer, $\mathcal{T^{\pi}}$ is the Bellman operator and $\mathcal{T^{\pi}}Q_{\theta}(s, a) = r(s, a) + \gamma \mathop{\mathbb{E}}_{a^{\prime} \sim \pi(\cdot|s^{\prime})} \left[ Q_{\overline{\theta}}(s^{\prime}, a^{\prime}) \right]$.



In offline RL, offline dataset $\mathcal{D}_{\text{off}} = \{(s_i, a_i, r_i, s_{i}^{\prime})\}_{i=1}^{|\mathcal{D}_{\text{off}}|}$ consists of transitions collected from the real environment $\mathcal{M}$ by some unknown behavior policy $\beta$.
Here, we additionally consider an {\em inaccurate simulator} modeled as an MDP $\widehat{\mathcal{M}} = (\mathcal{S, A}, \widehat{\mathcal{P}}, \rho_0, r, \gamma )$, where the dynamics of the inaccurate simulator $\widehat{\mathcal{P}}$ is different from the dynamics $\mathcal{P}$ in real environment. We can interact with such an inaccurate simulator by policy $\pi$ to collect the data $\mathcal{D}_\text{sim} = \{(s_i, a_i, r_i, s_{i}^{\prime})\}_{i=1}^{|\mathcal{D}_\text{sim}|}$. 
Our goal is to leverage the augmented dataset $\mathcal{D} = \mathcal{D}_{\text{off}} \cup \mathcal{D}_{\text{sim}}$ to learn a policy for completing tasks in the real environment.

Given a {\em limited offline dataset} and an {\em inaccurate simulator}, there are two key challenges: 1) exploitation-exploration trade-off when interacting with the simulator to collect useful data for learning and 2) reducing side-effect when leveraging both the real data and inaccurate simulation data.


\section{Method}

\begin{figure}[tbp]
    \centering
    {\includegraphics[width=1.0\columnwidth]{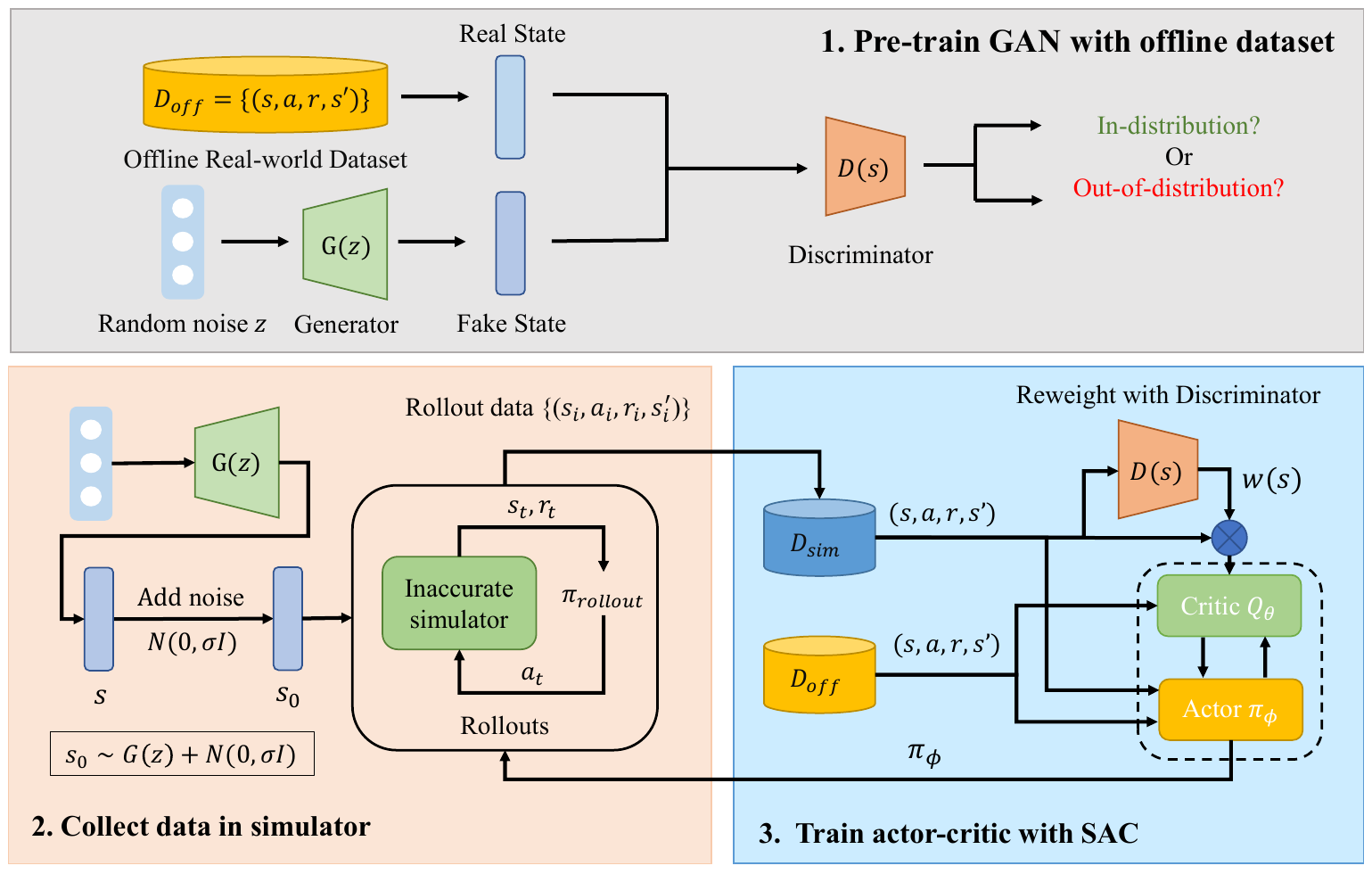}}
    \caption{Overview of our ORIS framework.}
    \label{Framework}
\end{figure}

\noindent Here, we propose our method, named \textbf{O}ffline \textbf{R}einforcement Learning with \textbf{I}naccurate \textbf{S}imulator (ORIS). 
As shown in Fig. \ref{Framework}, the GAN generator $G$ and discriminator $D$ are trained for addressing the aforementioned challenges. Specifically, the generator aims to generate states that align with the state distribution of the offline dataset, while the discriminator is tasked with discerning whether a given state is either in-distribution or out-of-distribution.
After that, we iteratively collect data using the inaccurate simulator and train the policy with the mixed dataset. 
During data collection, we interact with the simulator from the restart distribution provided by the generator $G$ to avoid aimless exploration. For the trade-off of exploitation-exploration, the rollout policy $\pi_{rollout}$ is a hybrid policy consisting of the random policy and the current policy $\pi_{\phi}$. 
During policy training, we utilize an off-policy actor-critic algorithm, i.e., SAC \cite{haarnoja2018soft}, training with both the offline dataset and the collected simulated data. Here, we adopt the weighted critic loss with discriminator $D$ to down-weight potentially harmful simulated data. Next, we will describe our method in more details.


\subsection{Pre-train GAN with Offline Dataset}
\noindent To improve offline RL with inaccurate simulator, the key issues are the exploitation-exploration trade-off when sampling and the inaccurate Q value estimation when minimizing the critic loss due to the dynamics gap. To avoid aimless exploration, we can sample trajectories starting from the state distribution fitting the offline dataset. To alleviate inaccurate Q value estimation, we need to discriminate potentially harmful data.
We will elaborate on these issues later. 

Fortunately, the structure of the generator and discriminator of GAN provides a promising solution to solve these two problems in a unified manner. Note that we choose GAN for its simplicity and effectiveness though any generative model with generator and discriminator (e.g., WGAN \cite{arjovsky2017wasserstein}) is compatible with our framework.
As shown in Fig. \ref{Framework}, the generator $G$ learns to produce samples that closely resemble the state distribution of the offline dataset, while the discriminator $D$ learns to distinguish between samples generated by $G$ and the real samples from the offline dataset:
\begin{equation}
    \min_{G}\max_{D}\mathop{\mathbb{E}}_{s \sim \mathcal{D}}\left[ \log (D(s))\right] + \mathop{\mathbb{E}}_{z \sim p(z)} \left[\log(1-D(G(z))) \right]
\end{equation}

To solve the min-max optimization problem, we update the generator and the discriminator iteratively. At every iteration $k$, we update $G_{\psi}^{k}$ and $D_{\omega}^{k}$ as follow:
\begin{small}
\begin{equation}
    G_{\psi}^{k+1} \leftarrow \mathop{\arg\min}\limits_{\psi} \mathop{\mathbb{E}}_{z \sim p(z)} \left[\log(1-D_{\omega}^{k}(G_{\psi}^{k}(z))) \right] 
    \label{generator loss}
\end{equation}
\end{small}
\begin{small}
\begin{equation}
    D_{\omega}^{k+1} \!\! \leftarrow \! \mathop{\arg\max}\limits_{\omega}\! \mathop{\mathbb{E}}_{s \sim \mathcal{D}}\!\left[ \log (D_{\omega}^{k}(s))\right]  + \! \!\!\!\!  \mathop{\mathbb{E}}_{z \sim p(z)}\! \left[\log(1\!-\!D_{\omega}^{k}(G_{\psi}^{k}(z))) \right] 
    \label{discriminator loss}
\end{equation}
\end{small}

After the GAN is trained, we use the generator $G_{\psi}$ to produce initial state distributions for the simulated process. Additionally, the discriminator $D_{\omega}$ is employed to assess the simulated data, in order to reweight it during the off-policy learning process. Next, we will illustrate how we use the generator and discriminator to better leverage the potential of the inaccurate simulator.

\subsection{Collecting Simulated Data with Generator} \label{Collecting Simulated Data with Generator}
\noindent As aforementioned, using the off-policy algorithm directly with limited offline data can lead to extrapolation errors stemming from the out-of-distribution (OOD) issue. This error cannot be fixed during offline training without additional online interaction. Therefore, our goal is to supplement the offline dataset with data from a simulator. However, it is still challenging to collect useful data for policy training due to the exploitation-exploration dilemma.

Most methods directly reset to initial states $\rho_0$ when sampling trajectories. However, this will lead to massive aimless exploration around $\rho_0$, especially for some robotic tasks with sparse reward. Fortunately, offline datasets usually consist of some trajectories completing the target task, which can provide valuable guidance for avoiding such exploration. 
For exploitation of the offline dataset, we utilize the generator of the pre-trained GAN model, which fits the state distribution of offline datasets. Specifically, we use $G(z)$ to generate the restart distribution $\rho_G \sim G(z) + \mathcal{N}(0, \sigma I)$. 
When interacting with the simulator starting from $s_0 \sim \rho_G$, it is more likely to sample high-return trajectories, which can be difficult to explore when starting from $\rho_0$.
For exploration within the simulator, we utilize a hybrid rollout policy similar to $\varepsilon$-greedy. Specifically, we choose the random policy with probability $p$, and the current policy with probability $1-p$. 
Such a hybrid rollout policy can explore the action space, effectively rectifying the extrapolation errors associated with OOD actions and thereby facilitating the discovery of an improved policy around the offline dataset. 
All in all, the restart distribution and the hybrid policy balance the exploitation-exploration dilemma in a more effective way.

\subsection{Reweighting Data with Discriminator} \label{Reweighting Data with Discriminator}

\noindent Although simulated data can complement limited offline data and alleviate unrealistic overestimation of the Q-value, treating all simulated and real data equally may harm the performance of the policy due to the dynamics gap. 
The primary dynamics gap between $\mathcal{M}$ and $\widehat{\mathcal{M}}$ originates from the environment dynamics $\mathcal{P}$, that is, $(s, a, r, s^{\prime}) \in \mathcal{D}_{\text{off}}$ and $(s, a, r, \widehat{s^{\prime}}) \in \mathcal{D}_{\text{sim}}$. Based on the Bellman equation, the Q-value estimates of similar state-action pairs $(s, a)$ are given by: 
$ Q_{\mathcal{M}}(s, a) = r + \gamma \mathop{\mathbb{E}}_{a^{\prime} \sim \pi(\cdot \mid s^{\prime})} \left[ Q(s^{\prime}, a^{\prime}) \right] $
and 
$ Q_{\widehat{\mathcal{M}}}(s, a) = r + \gamma \mathop{\mathbb{E}}_{a^{\prime} \sim \pi(\cdot \mid \widehat{s^{\prime}})} \left[Q(\widehat{s^{\prime}}, a^{\prime}) \right]$.
If we directly minimize the critic loss as delineated in Eq.\ref{critic loss}, the variance in the target Q-value, i.e., $Q_{\mathcal{M}}(s, a)$ and $Q_{\widehat{\mathcal{M}}}(s, a)$, may result in an ambiguous estimation of $Q(s, a)$. To address this, we propose to assign a lower weight to the transition in $\mathcal{D}_\text{sim}$. By doing so, we can reduce the impact on the overall Q-value estimation, and thereby alleviate the potential negative effects caused by the dynamics gap between $\mathcal{M}$ and $\widehat{\mathcal{M}}$.




Specifically, we employ adaptive weighting for simulated data to down-weight any potentially harmful simulated information. The corresponding weighted critic loss is as:
\begin{equation}
    \label{weighted critic loss}
    \begin{split}
        \mathcal{L}_{Q}(\theta) &= \! \mathop{\mathbb{E}}_{(s, a, s^{\prime}) \sim \mathcal{D}_{\text{off}}} \left[ (Q_{\theta}(s, a) - \mathcal{T^{\pi_{\phi}}}Q_{\theta}(s, a) )^{2} \right] \\
        & + \! \mathop{\mathbb{E}}_{(s, a, s^{\prime}) \sim \mathcal{D}_{\text{sim}}} \!w(s)\!\left[ (Q_{\theta}(s, a) - \mathcal{T^{\pi_{\phi}}}Q_{\theta}(s, a) )^{2} \right] 
    \end{split}
\end{equation}
where $w(s) = \mathop{\text{clip}}(1 - 2D(s), w_{\text{min}}, w_{\text{max}})$, and $w_{\text{min}}$ is a small positive number. The underlying idea behind the weight $w(s)$ is as follows. For a given transition $(s_{\text{sim}},a_{\text{sim}},r_{\text{sim}},s_{\text{sim}}^{\prime}) \in \mathcal{D}_{\text{sim}}$, if the state $s_{\text{sim}}$ is in-distribution (i.e., the offline dataset already contains transitions of $s_{\text{sim}}$), we assign a small weight to that transition when calculating the critic loss. This prevents inconsistency in Q-value estimation for $s_{\text{sim}}$ between the offline data and simulated data. In such cases, $D(s) \rightarrow 0.5$ and the weight $w(s) \rightarrow w_{\text{min}}$.
Conversely, if the state $s_{\text{sim}}$ is out-of-distribution and absent from the offline dataset, there is no need to assign a small weight to it, as it will not conflict with offline data.
For such states, $D(s) \rightarrow 0$ and the weight $w(s) \rightarrow w_{\text{max}}$.

\begin{algorithm}[t]\small
    \caption{\textbf{O}ffline \textbf{R}L with \textbf{I}naccurate \textbf{S}imulator (ORIS)}
    \label{algo}
    \begin{algorithmic}[1]
        \REQUIRE offline dataset $\mathcal{D}_{\text{off}}$, inaccurate simulator with biased dynamics $\widehat{\mathcal{P}}$, rollout horizon $H$, rollout count $C$ 
        \ENSURE policy network $\pi_{\phi}$
        \STATE Initialize discriminator $D_{\omega}(\cdot | s)$ and generator $G_{\psi}(\cdot | z)$
        \STATE Initialize policy network $\pi_{\phi}$ and critic network $Q_{\theta}$
        \STATE Initialize replay buffer $\mathcal{D}_{\text{sim}} \leftarrow \emptyset$
        \STATE Pre-train $D_{\omega}(\cdot | s)$ and $G_{\psi}(\cdot | z)$ using Eq.\ref{generator loss} and Eq.\ref{discriminator loss}
        
        
        \FOR{$epoch = 0, 1, \cdots$}
            \STATE $\pi_{\text{rollout}} =  \begin{cases} 
                    \pi_{\text{random}} & \text{with probability}\ p \\
                    \pi_{\phi} & \text{with probability}\ 1-p
                    \end{cases}$
            \FOR[Collect data in simulator]{$c = 0, 1, \dots, C$}
                \STATE Generate a restart state $s_0 \sim G(z) + \mathcal{N}(0, \sigma I)$
                \STATE Rollout $H$ steps in the simulator starting from $s_0$ by $\pi_{\text{rollout}}$, and add the rollout to $\mathcal{D}_{\text{sim}}$
            \ENDFOR
            \STATE Sample minibatch data $B_{\text{off}} \sim \mathcal{D}_{\text{off}}, B_{\text{sim}} \sim \mathcal{D}_{\text{sim}}$
            \STATE Update $Q_{\theta}$ and $\pi_{\phi}$ via minimizing Eq.\ref{weighted critic loss} and Eq.\ref{new actor loss}
        \ENDFOR
    \end{algorithmic}
\end{algorithm}

Building upon this, we minimize the actor loss by directly minimizing it over the dataset combining both the offline and simulated data, $\mathcal{D}_{\text{off}} \cup \mathcal{D}_{\text{sim}}$. The updated actor loss is as:
\begin{equation}
    \label{new actor loss}
    \mathcal{L}_{\pi}(\phi) = \mathop{\mathbb{E}}_{\substack{s \sim \mathcal{D}_{\text{off}} \cup \mathcal{D}_{\text{sim}}, \\ a \sim \pi_{\phi}(\cdot|s)}} \left[- Q_{\theta}(s, a) + \lambda \log{\pi_{\phi}(a|s)} \right]
\end{equation}
where $\lambda$ is the temperature parameter as in SAC. 
Note that ORIS can be compatible with any off-policy actor-critic algorithm (e.g., TD3 \cite{fujimoto2018addressing}). 
With the components described above, the main procedures of ORIS are outlined in Algorithm \ref{algo}. 
\begin{table*}[ht]
    \caption{Average normalized scores of all methods. 
    (g2.0) means the simulator with 2 times gravity.}
    \label{tab:result_offline}
    \centering 
    \resizebox{0.95\textwidth}{!}{
    \begin{tabular}{ l | c c c c c c | c c c}
        \hline
        {\bf Task Name}  & {\bf BC} & {\bf CQL} & {\bf IQL} & {\bf TD3+BC} & {\bf COMBO} & {\bf SQL} & {\bf SAC(g2.0)} & {\bf H2O(g2.0)} &  {\bf ORIS(g2.0)(Ours)}  \\ 
        \hline
            hopper-random             & 3.7$\pm$0.6   & 7.9$\pm$0.4     & 7.9$\pm$0.2    & 8.5$\pm$0.6    & 17.9$\pm$1.4   & 7.8$\pm$0.5   & 10.0$\pm$5.6      & 20.6$\pm$9.8  & \bf{31.4$\pm$0.4}    \\ [2pt]
            hopper-medium-replay      & 16.6$\pm$4.8  & 88.7$\pm$12.9   & 94.7$\pm$8.6   & 60.9$\pm$18.8  & 89.5$\pm$1.8   & 96.7$\pm$3.3  & 10.0$\pm$5.6      & 46.7$\pm$28.0 & \bf{100.6$\pm$0.3}   \\ [2pt]
            hopper-medium             & 54.1$\pm$3.8  & 53.0$\pm$28.5   & 66.2$\pm$5.7   & 59.3$\pm$4.2   & 97.2$\pm$2.2   & 73.5$\pm$3.4  & 10.0$\pm$5.6      & 21.6$\pm$17.7 & \bf{99.8$\pm$0.8}    \\ [2pt]
            hopper-medium-expert      & 53.9$\pm$4.7  & 105.6$\pm$12.9  & 91.5$\pm$14.3  & 98.0$\pm$9.4   & 111.1$\pm$2.9  & \bf{111.8$\pm$2.2} &  10.0$\pm$5.6   & 25.2$\pm$24.4  & 110.1$\pm$1.5    \\ [2pt]
            \hline 
            walker2d-random           & 1.3$\pm$0.1   & 5.1$\pm$1.3     & 5.4$\pm$1.2    & 1.6$\pm$1.7    & 7.0$\pm$3.6    & 5.1$\pm$0.4   & 30.2$\pm$19.9     & 12.1$\pm$6.3  & \bf{30.4$\pm$15.0}   \\ [2pt]
            walker2d-medium-replay    & 20.3$\pm$9.8  & 81.8$\pm$2.7    & 73.8$\pm$7.1   & 81.8$\pm$5.5   & 56.0$\pm$8.6   & 77.2$\pm$3.8  & 30.2$\pm$19.9     & 39.2$\pm$25.9 & \bf{91.5$\pm$0.5}    \\ [2pt]
            walker2d-medium           & 70.9$\pm$11.0 & 73.3$\pm$17.7   & 78.3$\pm$8.7   & 83.7$\pm$2.1   & 81.9$\pm$2.8   & 84.2$\pm$4.6  & 30.2$\pm$19.9     & 34.4$\pm$15.2 & \bf{86.2$\pm$5.3}    \\ [2pt]
            walker2d-medium-expert    & 90.1$\pm$13.2 & 107.9$\pm$1.6   & 109.6$\pm$1.0  & \bf{110.1$\pm$0.5}  & 103.3$\pm$5.6   & 110.0$\pm$0.8 &  30.2$\pm$19.9  & 27.3$\pm$17.0  & 102.8$\pm$2.2   \\ [2pt]
            \hline
            halfcheetah-random        & 2.2$\pm$0.0   & 17.5$\pm$1.5    & 13.1$\pm$1.3   & 11.0$\pm$1.1   & 38.8$\pm$3.7   & 14.4$\pm$1.0  & \bf{43.3$\pm$1.5} & 35.2$\pm$1.4  & 39.2$\pm$1.7         \\ [2pt]
            halfcheetah-medium-replay & 37.6$\pm$2.1  & 45.5$\pm$0.7    & 44.2$\pm$1.2   & 44.6$\pm$0.5   & 55.1$\pm$1.0   & 44.8$\pm$0.7  & 43.3$\pm$1.5      & 52.8$\pm$5.5  & \bf{59.6$\pm$2.4}    \\ [2pt]
            halfcheetah-medium        & 43.2$\pm$0.6  & 47.0$\pm$0.5    & 47.4$\pm$0.2   & 48.3$\pm$0.3   & 54.2$\pm$1.5   & 48.3$\pm$0.2  & 43.3$\pm$1.5      & 55.2$\pm$4.9  & \bf{68.2$\pm$2.1}    \\ [2pt]
            halfcheetah-medium-expert & 44.0$\pm$1.6  & 75.6$\pm$25.7   & 86.7$\pm$5.3   & 90.7$\pm$4.3   & 90.0$\pm$5.6   & \bf{94.0$\pm$0.4}  & 43.3$\pm$1.5 &   33.0$\pm$6.6  & 74.5$\pm$4.9       \\ [2pt]
            \hline
            {\bf Average Score} & 36.5$\pm$4.4 & 59.1$\pm$8.9 & 59.9$\pm$4.6 & 58.2$\pm$4.1 & 66.8$\pm$3.4 & 64.0$\pm$1.8 & 27.8$\pm$9.0 & 33.6$\pm$13.6 & \bf{74.5$\pm$3.1} \\ [2pt]
          \hline
    \end{tabular}
    }
\end{table*}

\section{Experiments}

\noindent We pose the following questions and provide affirmative answers in our experiments:
Q1) Is ORIS effective for RL with offline data and an inaccurate simulator?
Q2) When the offline data is reduced, can ORIS make full use of the simulator to supplement the data?
Q3) Is ORIS robust to the inaccuracy of the simulator?
Q4) How do the different components affect the performance?
Q5) Is ORIS effective in real-world robotic tasks?

\subsection{D4RL Benchmarks and Baselines}
\noindent We first evaluate our method on the locomotion tasks of the widely used D4RL \cite{fu2020d4rl} dataset, i.e., \textit{halfcheetah}, \textit{hopper} and \textit{walker2d}, and use the MuJoCo physics simulator \cite{todorov2012mujoco}. 
For each domain, we reconstruct three task simulation environments with intentionally introduced dynamics gaps upon the original locomotion tasks (which serve as the real environments) by modifying the dynamics parameters identical to H2O \cite{niu2022trust}: 1) \textbf{Gravity}: applying 2 times the gravitational acceleration in the simulator; 2) \textbf{Friction}: using 0.3 times the friction coefficient; 3) \textbf{Action Noise}: adding a random noise sampled from a standard normal distribution $\mathcal N(0,1)$ on every dimension of the action space. 

We compare our method with the state-of-the-art online, offline, and hybrid offline-and-online RL methods. For the online RL, we compare with SAC \cite{haarnoja2018soft}, which is trained in the modified simulator, and evaluate the policy in the original environment. For the offline RL, we compare with behavior cloning (BC), CQL \cite{kumar2020conservative}, IQL \cite{kostrikov2021offline}, TD3+BC \cite{fujimoto2021minimalist}, COMBO \cite{yu2021combo}, SQL \cite{xu2023offline}, which are trained on the fixed offline dataset. The results of offline baselines are taken directly from their corresponding papers. For the hybrid method, we compare with H2O \cite{niu2022trust}, which uses the same settings as ours. 
During training, we collect simulated data from the modified (inaccurate) simulator and evaluate the learned policy on the original (accurate) simulator. 
The policy is trained for 500K steps and evaluated over 10 episodes every 1000 steps. The results are over five random seeds.

\subsection{Results on D4RL Benchmarks}\label{exp_text_main}

\noindent To answer question Q1, we compare ORIS with all the baselines on the D4RL dataset. 
In Table \ref{tab:result_offline}, SAC, H2O, and ORIS collect simulated data from the simulator with 2 times the gravitational acceleration, i.e., the (g2.0) suffix. ORIS outperforms the baselines in most of the tasks. 
Offline RL performs well on \textit{medium-expert} datasets that contain expert trajectories. However, they perform poorly on other types of datasets, which usually contain sub-optimal human demonstrations. The results show that the performance of offline RL is greatly affected by the quality of the dataset.
H2O behaves relatively well in \textit{halfcheetah} but poorly in \textit{hopper} and \textit{walker2d} tasks because the agent quickly falls down and is unable to explore valuable data in the simulator when starting from $\rho_0$.
In contrast, ORIS starts the rollout from the states generated by GAN and achieves good performance, which benefits the balance of exploitation of the offline data and exploration of the inaccurate simulator. 

In addition, we compare SAC and H2O under different simulators with varying dynamics, and the results are shown in Table \ref{tab:result_online}. It can be seen that ORIS consistently outperforms H2O, which demonstrates that our method can take more advantages than H2O from the inaccurate simulator. SAC suffers from the dynamics gap in most cases but sometimes outperforms ORIS on random datasets. This is because our method trusts the offline dataset, which may be misleading especially in the random cases.

\begin{table*}[ht]
    \caption{Average normalized scores of SAC, H2O and our method for simulators with different unreal dynamics. }
    \label{tab:result_online}
    \centering
    \resizebox{\textwidth}{!}{
        \begin{tabular}{ l | c c c | c c c | c c c}
            \hline
            {\bf Unreal Dynamics} & \multicolumn{3}{|c|}{\bf Gravity} & \multicolumn{3}{|c|}{\bf Friction}& \multicolumn{3}{|c}{\bf Action Noise}  \\
            \hline
            {\bf Task Name}             & {\bf SAC}    & {\bf H2O}    & {\bf ORIS(Ours)}    & {\bf SAC}     &  {\bf H2O}    & {\bf ORIS(Ours)}    & {\bf SAC}    & {\bf H2O}     & {\bf ORIS(Ours)}    \\ 
            \hline
            hopper-random               & 10.0$\pm$5.6 & 20.6$\pm$9.8 & \bf{31.4$\pm$0.4}   & \bf{56.8$\pm$31.8} & 15.8$\pm$7.3  & 34.2$\pm$7.7  & \bf{37.4$\pm$18.2} & 13.7$\pm$3.2  & 34.4$\pm$14.4 \\ [2pt]
            hopper-medium-replay        & 10.0$\pm$5.6 & 46.7$\pm$28.0& \bf{100.6$\pm$0.3}  & 56.8$\pm$31.8 & 43.0$\pm$28.6 & \bf{102.0$\pm$1.3}  & 37.4$\pm$18.2 & 66.3$\pm$28.7 & \bf{101.6$\pm$1.1} \\ [2pt]
            hopper-medium               & 10.0$\pm$5.6 & 21.6$\pm$17.7& \bf{99.8$\pm$0.8}  & 56.8$\pm$31.8 & 24.9$\pm$9.7  & \bf{101.1$\pm$0.8} & 37.4$\pm$18.2 & 75.2$\pm$16.5  & \bf{98.5$\pm$3.2} \\ [2pt]
            hopper-medium-expert        & 10.0$\pm$5.6 & 25.2$\pm$24.4& \bf{110.1$\pm$1.5}  & 56.8$\pm$31.8 & 15.7$\pm$14.6 & \bf{106.2$\pm$7.7} & 37.4$\pm$18.2 & 31.4$\pm$20.9 & \bf{109.1$\pm$3.9}  \\ [2pt]
            \hline
            walker2d-random             & 30.2$\pm$19.9 & 12.1$\pm$6.3 & \bf{30.4$\pm$15.0} & \bf{76.1$\pm$8.9}  & 9.3$\pm$5.2   & 22.5$\pm$16.8 & \bf{19.7$\pm$7.8} & 11.4$\pm$5.5  & 16.6$\pm$4.0    \\ [2pt]
            walker2d-medium-replay      & 30.2$\pm$19.9 & 39.2$\pm$25.9 & \bf{91.5$\pm$0.5} & 76.1$\pm$8.9  & 75.6$\pm$12.7  & \bf{93.5$\pm$9.9} & 19.7$\pm$7.8 & 36.4$\pm$18.6 & \bf{95.3$\pm$3.6}   \\ [2pt]
            walker2d-medium             & 30.2$\pm$19.9 & 34.4$\pm$15.2& \bf{86.2$\pm$5.3}  & 76.1$\pm$8.9  & 30.4$\pm$9.3  & \bf{89.6$\pm$9.9}  & 19.7$\pm$7.8 & 30.4$\pm$6.9  & \bf{88.0$\pm$5.7}   \\ [2pt]
            walker2d-medium-expert      & 30.2$\pm$19.9 & 27.3$\pm$17.0 & \bf{102.8$\pm$2.2} & 76.1$\pm$8.9 & 44.7$\pm$14.4 & \bf{103.4$\pm$1.8} & 19.7$\pm$7.8 & 40.4$\pm$13.7 &  \bf{107.7$\pm$1.1} \\ [2pt]
            \hline
            halfcheetah-random          & \bf{43.3$\pm$1.5} & 35.2$\pm$1.4 & 39.2$\pm$1.7   & 41.8$\pm$3.7  & 42.7$\pm$15.6 & \bf{53.0$\pm$4.2}  & \bf{41.2$\pm$3.8} & 9.3$\pm$0.4    & 20.5$\pm$0.3  \\ [2pt]
            halfcheetah-medium-replay   & 43.3$\pm$1.5 & 52.8$\pm$5.5 & \bf{59.6$\pm$2.4}   & 41.8$\pm$3.7  & 53.9$\pm$4.9  & \bf{65.6$\pm$3.0}  & 41.2$\pm$3.8 & 53.9$\pm$4.2   & \bf{59.6$\pm$0.8}  \\ [2pt]
            halfcheetah-medium          & 43.3$\pm$1.5 & 55.2$\pm$4.9 & \bf{68.2$\pm$2.1}   & 41.8$\pm$3.7  & 51.1$\pm$7.0  & \bf{73.2$\pm$2.3}  & 41.2$\pm$3.8 & 60.1$\pm$2.7   & \bf{66.2$\pm$1.3}  \\ [2pt]
            halfcheetah-medium-expert   & 43.3$\pm$1.5 & 33.0$\pm$6.6 & \bf{74.5$\pm$4.9}   & 41.8$\pm$3.7  & 18.4$\pm$6.2  & \bf{86.8$\pm$3.5}  & 41.2$\pm$3.8 & 33.7$\pm$8.6 & \bf{76.0$\pm$7.8}   \\ [2pt]
            \hline
            {\bf Average Score} & 27.8$\pm$9.0 & 33.6$\pm$13.6 & \bf{74.5$\pm$3.1} & 58.2$\pm$14.8 & 35.5$\pm$11.3 & \bf{77.6$\pm$5.7} & 32.8$\pm$9.9 & 38.5$\pm$10.8 & \bf{72.8$\pm$3.9} \\ [2pt]
            \hline
        \end{tabular}
    }
\end{table*}


\subsection{Results on Small Dataset}

\noindent 
\noindent Given a small offline dataset, we aim to assess if our method can effectively leverage the simulator to supplement the data. For the limited dataset, we split the \textit{hopper-medium-replay-v2} dataset into trajectories and then randomly selected subsets comprising 25\% and 5\% of these trajectories to create two new limited datasets. We compare our method with offline baselines and H2O given the limited datasets. 

As shown in Fig. \ref{fig: Limited}, the performances of almost all offline RL methods drop dramatically as the amount of data decreases, which shows that offline RL methods are extremely sensitive to data quantity. 
H2O also performs worse with less data, showing that the performance of H2O also highly relies on the amount of offline data. 

By leveraging the simulator, our method outperforms all baselines on all limited datasets. Remarkably, the final scores only dropped 8\% when the amount of data decreased from 100\% to 5\%. This demonstrates that our method can effectively use the simulator to compensate for the extremely little data. We attribute this robust performance to the generator's ability to accurately fit the state distribution even with limited offline datasets. Furthermore, the simulated data, collected from restart distribution $\rho_{G}$, proves to be a more effective supplement to the limited offline data.

\subsection{Results on Robustness to Simulator Inaccuracy}

\noindent In the subsection, we investigate the robustness of our method with varying simulators. In complex applications, the gap between the simulator and the real environment can be substantial. 
Specifically, we compare our method with SAC and H2O on the \textit{hopper-medium-replay-v2} while increasing the simulator's inaccuracy by applying a scaling factor of \textit{Gravity Coefficient (GC)} to the gravitational acceleration in the simulator, i.e., $\textit{GC}=3, 4, 5$. Intuitively, the inaccuracy of the simulator grows as \textit{GC} increases. 

\begin{figure}[tbp]
    \centering
    \subfigure[Small Datasets] 
    {
        \includegraphics[width=.45\columnwidth]{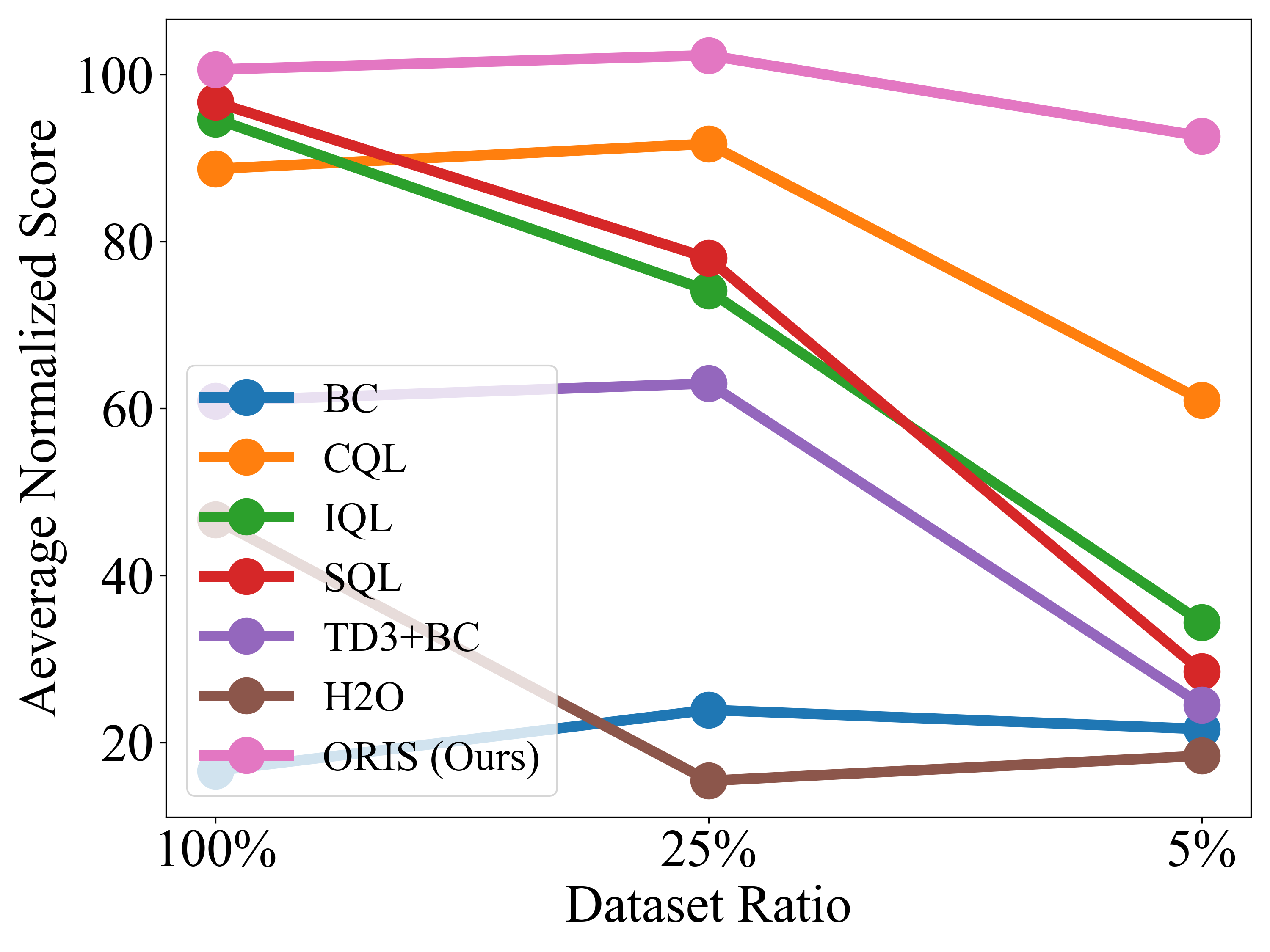}
        \label{fig: Limited}
    }
    \subfigure[Simulator Inaccuracies] 
    {
        \label{fig: Robustness}
        \includegraphics[width=.45\columnwidth]{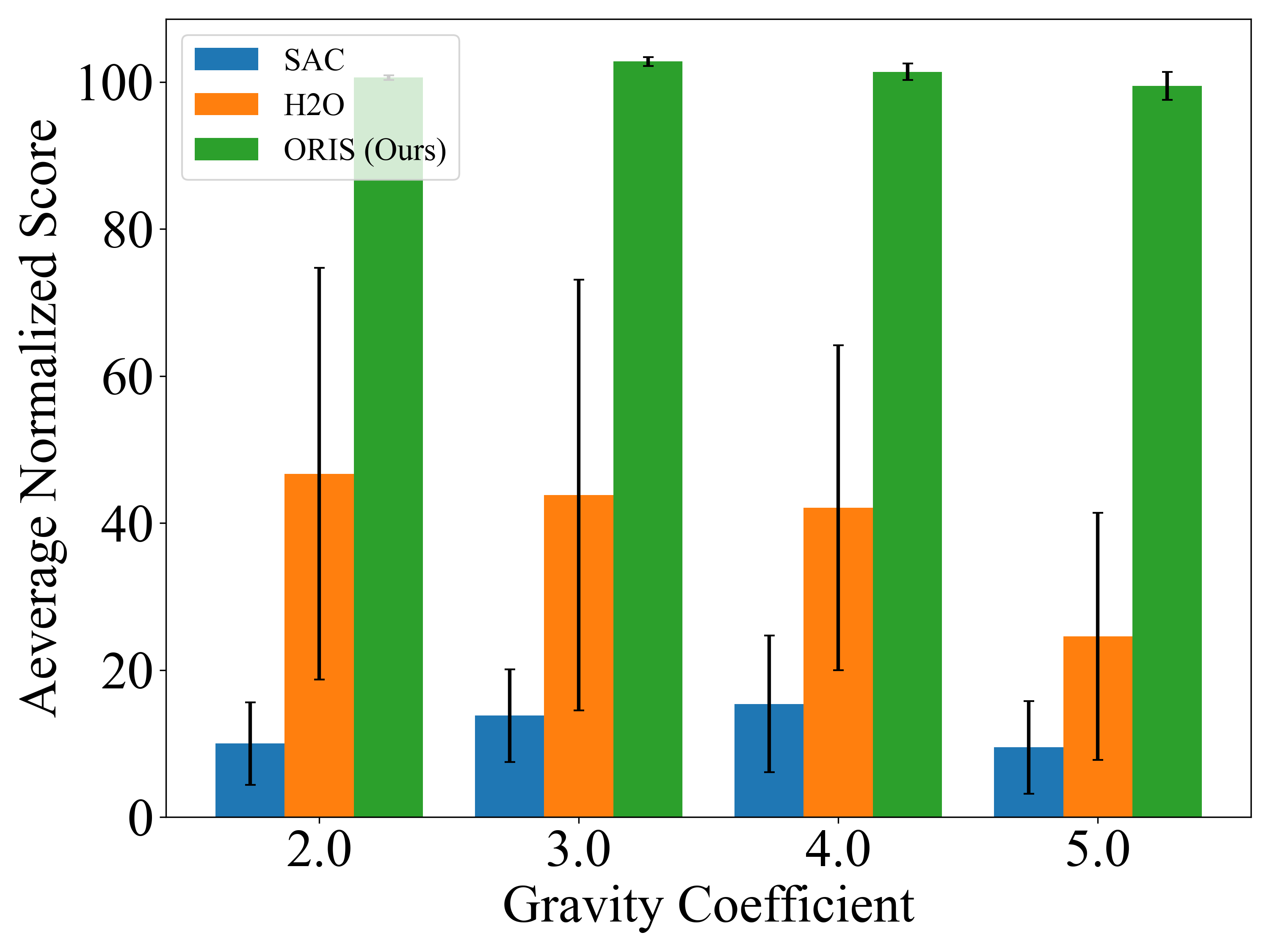}
    }
    
    \caption{Results on small datasets (25\% and 5\%) (left) and different simulator inaccuracies (right).}
\end{figure}

As shown in Fig. \ref{fig: Robustness}, the performance of H2O drops when \textit{GC} increases, which is because H2O cannot fully utilize the simulator with a large gap and accurately estimate the value function of state-action pair from offline dataset due to the poor coverage of state-action space.
Without the information of the real dynamics, SAC is severely affected by the dynamics gap, and cannot repair the impact of the gap. Our method is robust to the gap and the performance only drops slightly when $\textit{GC}=5$. These results show that our method can better utilize the inaccurate simulator and is more robust to inaccurate factors.

\subsection{Ablation Studies}

\begin{figure}[t]
	\centering

	{\includegraphics[width=.7\columnwidth]{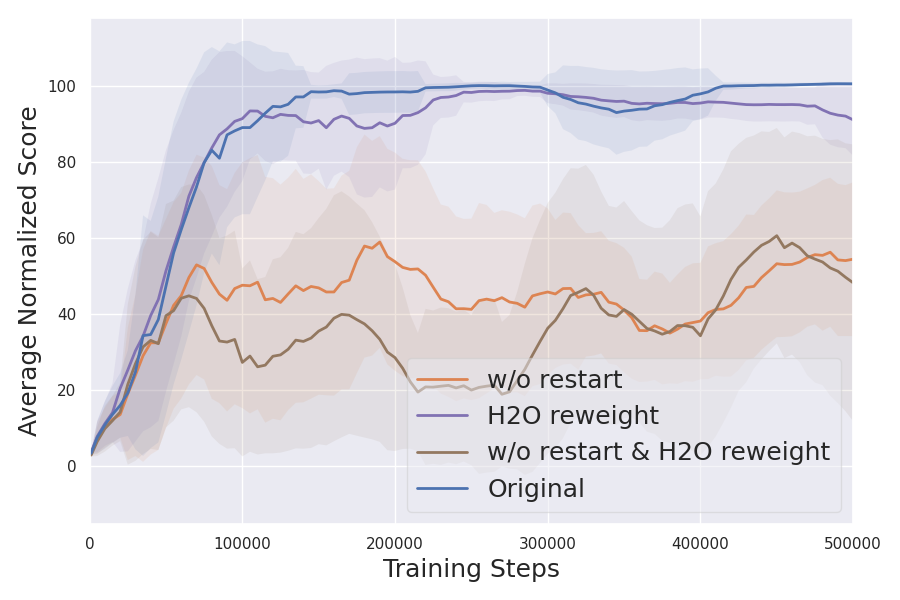}}
 	\caption{Ablation experiments for different modules.}
	\label{ablation_exp}
\end{figure}

\noindent In the subsection, we verify the effectiveness of the components of our approach on \textit{hopper-medium-replay-v2} datasets. 
We test 1) \textbf{w/o restart}: starting from the initial state $s_0 \sim \rho_0$ to replace the distribution generated by the GAN generator in our method and keep the rest of the modules unchanged; 2) \textbf{H2O reweight} $w(s, a, s')$ (Eq. 8 in \cite{niu2022trust}) for simulated data instead of our discriminator-based re-weighting; 3) \textbf{w/o restart \& H2O reweight}: replace both start distribution and re-weighting method, which is the same as H2O.

As shown in Fig. \ref{ablation_exp}, the \textbf{Original} version of our method achieves the best performance. The performance of the \textbf{w/o restart} version drops by a significant margin (46.9\%), which shows that initiating from the states produced by the GAN generator can effectively enhance performance.
For the re-weighting module, our approach outperforms \textbf{H2O reweight} (with a drop of about 9.4\%). These results highlight the effectiveness of our re-weighting module.

\subsection{Real-World Robotic Manipulation}
\begin{figure}[t]
    \centering
    \subfigure[MuJoCo Metaworld Simulator]{
        \includegraphics[width=0.45\columnwidth]{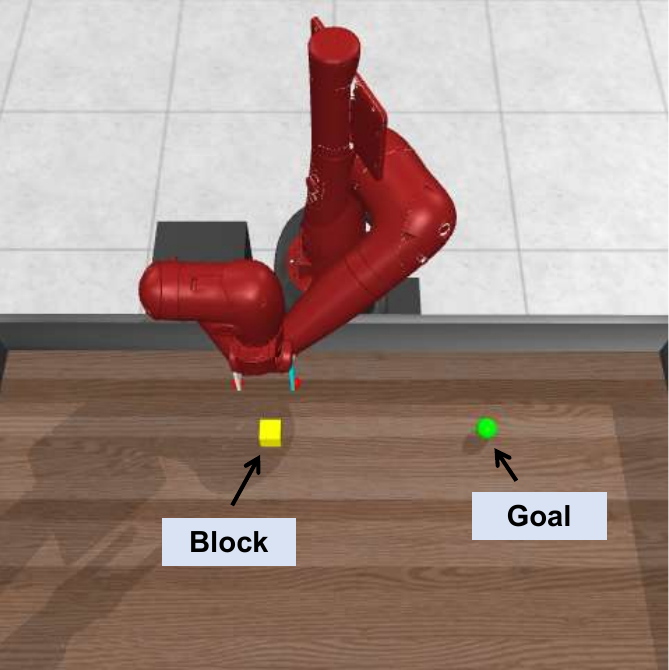}
    } 
    \subfigure[Real-world Kinova Arm]{
        \includegraphics[width=0.415
        \columnwidth]{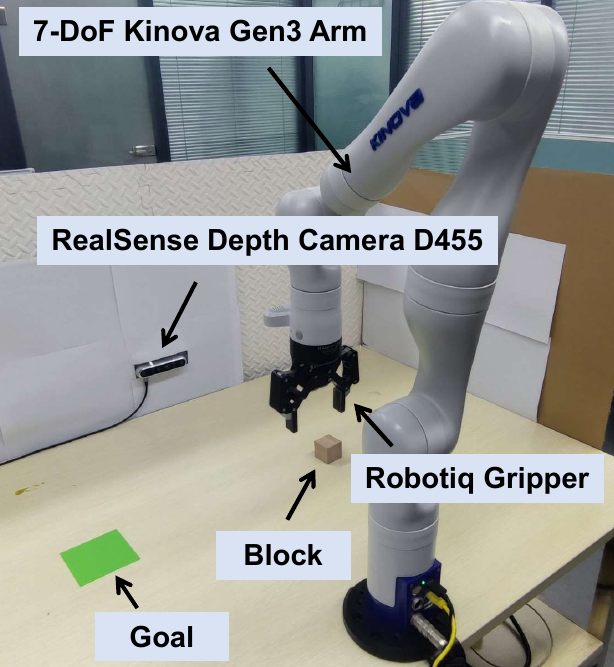}
    }
    \caption{Pick-and-place in simulation (left) and real (right).}
    \label{kinova}
\end{figure}

\noindent We now test the performance of ORIS to a real-world robot, which is typically more complex and involve many factors contributing to the simulation gap (e.g., modeling error, calibration error, control inaccuracy). 
As shown in Fig. \ref{kinova}, we use a 7-DoF Kinova Gen3 robotic arm equipped with a Robotiq gripper to do the \textit{Pick-and-place} task. 
The state space is represented by a 7-dimensional vector, including end-effector position (3D), gripper opening state (1D), and position of the block (3D). The block's pose is estimated using Apritag \cite{olson2011apriltag} from RGB images captured by the RealSense depth camera D455. The action space is a 4-dimensional vector, responsible for dictating the velocity of the end-effector (3D) and the status of the gripper (1D, i.e., open or close). The reward function is defined as $r=-0.1+0.1\times\mathbb{I}(d_1<0.1)+20.0\times\mathbb{I}(d_2<0.05)$, where $\mathbb{I}(\cdot)$ is the indicator function, $d_1$ is the distance between the gripper and the object and $d_2$ is the distance between the object and the goal. 
To collect an offline dataset, we employed a rule-based policy created manually, which generated trajectories for placing the object at various positions.
For the simulator, we use an off-the-shelf simulation environment: Metaworld \cite{yu2020meta}, based on Mujoco \cite{todorov2012mujoco}, without changing its robot model and parameters. 
During evaluation, the object is initially placed at 25 different positions in each episode. 

The results are presented in Table \ref{tab:kinova result}. Among offline RL methods trained solely on the offline dataset, CQL and TD3+BC both encounter severe OOD errors, rendering them unable to pick up the block.
While BC, IQL, and SQL can reach and grasp the object, they struggled to accurately place it in the correct position. This failure can be attributed to the low-quality nature of the dataset, which includes numerous task-agnostic trajectories. Among the methods that used the inaccurate simulator, SAC was able to successfully complete the task within the simulator. However, when transferring the learned policy to the real world, it faced challenges in both picking up and placing the object due to the dynamics gap.
H2O performed poorly as it did not explore sufficiently within the simulator, hindering its ability to learn effective policies. In contrast, ORIS can complete this challenging task due to the proper utilization of both the limited offline data and the inaccurate simulator, which shows the effectiveness of our method in the real-world applications.

\section{Related Work}
\noindent Offline RL aims to address the problem of learning effective policies entirely from previously collected data using some behavior policies, without further online interaction \cite{fujimoto2019off, levine2020offline}. The main challenge is the extrapolation error \cite{fujimoto2019off}, the overestimation of the value function of OOD actions, which is caused by the distribution shift between the current policy and the behavior policy \cite{kumar2020conservative, kumar2019stabilizing}. Several methods such as BCQ \cite{fujimoto2019off}, BEAR \cite{kumar2019stabilizing}, BRAC \cite{wu2019behavior}, TD3+BC \cite{fujimoto2021minimalist}, and LAPO \cite{chen2022lapo} constrain the learned policy to the behavior policy used to collect the dataset. Other methods such as CQL \cite{kumar2020conservative} and COMBO \cite{yu2021combo} constrain the learned policy by making conservative estimates of value functions of OOD actions. 
Another type of methods such as AWR \cite{peng2019advantage}, One-step \cite{brandfonbrener2021offline}, IQL \cite{kostrikov2021offline} and SQL \cite{xu2023offline} does behavior cloning weighted by advantage of the data. 
Offline RL methods are often conservative and pessimistic, and their performance heavily depends on the quality, size, and coverage of the state-action space of the given offline dataset \cite{arnob2021importance, chen2019information, schweighofer2022dataset}.

\begin{table}[t]
    \caption{Average return and success rate (SR) of all methods for the Pick-and-place task. The sub-tasks of ``Reach'', ``Pick'' and ``Place'' success when  $d_1$$<$0.1, the height of the object $z_{obj}$$>$0.05 and $d_2$$<$0.05 respectively.}
    \label{tab:kinova result}
    \centering
    \resizebox{0.49\textwidth}{!}{
        \begin{tabular}{ c | c c c c c c c c}
            \hline
            {\bf }   & {\bf BC} & {\bf CQL} & {\bf IQL} & {\bf TD3+BC} & {\bf SQL} & {\bf SAC} & {\bf H2O} & {\bf ORIS} \\ 
            \hline
            Average return & -6.50 & -13.58 & -3.84 & -14.99     & -3.26 & 4.20 & -14.56 & \bf{18.18} \\ [2pt]
            SR (Reach) & 98.7\% & 29.3\% & 100.0\% & 6.7\% & 98.7\% & 100.0\% & 22.7\% & \bf{100.0\%} \\ [2pt]
            SR (Pick) & 61.3\% & 0.0\% & 92.0\% & 0.0\% & 97.3\% & 30.3\% & 0.0\% & \bf{98.7\%} \\ [2pt]
            SR (Place) & 0.0\% & 0.0\% & 0.0\% & 0.0\% & 4.0\% & 27.6\% & 0.0\% & \bf{98.7\%} \\ [2pt]
            \hline
        \end{tabular}
    }
\end{table}

Another line of work does data augmentations to the offline dataset. S4RL \cite{sinha2022s4rl} adds noise to the states in the offline dataset. Some model-based methods augment the offline dataset by learning a reverse model \cite{wang2021offline} or bidirectional models \cite{lyu2022double}. \cite{jang2023k} propose mixup augmentation in the Koopman subspace for offline RL. Similar to ours, H2O \cite{niu2022trust} uses an inaccurate simulator to supplement the dataset, which learns a pair of discriminators to reweight the simulated data. 
However, H2O does not fully utilize the simulator. Our method uses the same settings with H2O but fully utilizes the simulator to augment the offline dataset by starting from state-distribution \cite{nair2018overcoming, sharma2022state} and re-weight simulated data according to the state-distribution. 

Our work is also related to sim-to-real transfer \cite{zhao2020sim}. Domain randomization randomizes the simulation to cover the real distribution of real-world data, including visual randomization \cite{tobin2017domain, tremblay2018training, matas2018sim} and dynamics randomization \cite{peng2018sim}. Some other works based on system identification \cite{kaspar2020sim2real}, attempt to build a precise simulator for sim-to-real transfer. Unlike these works, ORIS does not require access to the parameters of the simulator for adjustment and can directly utilize the off-the-shelf simulator for real-world tasks.

\section{Conclusions}

\noindent In this paper, we proposed a method ORIS to enhance the performance of offline RL by utilizing an inaccurate simulator.
Firstly, we train a GAN to fit the state distribution of the offline dataset. 
Then we use the generator to produce starting states for rollouts in the simulator with a hybrid behavior policy, aiming to balance the exploration-exploitation dilemma of collecting useful data.
Furthermore, we employ the discriminator to re-weight the simulated data for more precise Q value estimation.
Our experiments on the D4RL benchmarks and the real-world task demonstrate the effectiveness of our approach.
In the future, we plan to extend our method to tackle more challenging robotics tasks.

\bibliographystyle{IEEEtran} 
\bibliography{mybibfile}
\end{document}